# oRetrieval Augmented Generation for 10 Large Language Models and its Generalizability in Assessing Medical Fitness

Yu He Ke, MMed(Anes)*1,2
Liyuan Jin, MD*3,4,5
Kabilan Elangovan, BEng4,5
Hairil Rizal Abdullah, PhD1,2
Nan Liu, PhD3
Alex Tiong Heng Sia^, MMed(Anes)3,7
Chai Rick Soh, MMed(Anes)1,3
Joshua Yi Min Tung, MBBS2,6
Jasmine Chiat Ling Ong, PharmD3,8
Chang-Fu Kuo, 9, 10, 11
Shao-Chun Wu, 12, 13
Vesela P. Kovacheva, M.D. Ph.D. 14
Daniel Shu Wei Ting,PhD[+]3,4,5

Affiliations:
1 Department of Anesthesiology, Singapore General Hospital, Singapore, Singapore
2 Data Science and Artificial Intelligence Lab, Singapore General Hospital, Singapore
3 Duke-NUS Medical School, Singapore, Singapore
4 Singapore National Eye Centre, Singapore Eye Research Institute, Singapore, Singapore
5 Singapore Health Services, Artificial Intelligence Office, Singapore
6 Department of Urology, Singapore General Hospital, Singapore
7 Department of Women's Anaesthesia, KK Women's and Children's Hospital, Singapore.
8 Division of Pharmacy, Singapore General Hospital, Singapore
9 Department of Rheumatology,  Allergy, and Immunology, Chang Gung Memorial Hospital, 333, Taiwan.
10 Center for Artificial Intelligence in Medicine, Chang Gung Memorial Hospital, 333, Taiwan.
11 School of Medicine, Chang Gung University, Taoyuan, 333, Taiwan.
12 Department of Anesthesiology, Kaohsiung Chang Gung Memorial Hospital, College of Medicine, Chang Gung University, Kaohsiung 833, Taiwan.
13 School of Medicine, College of Medicine, National Sun Yat-sen University, Kaohsiung 804, Taiwan.
14 Department of Anesthesiology, Perioperative and Pain Medicine, Brigham and Women's Hospital, Harvard Medical School, Boston, MA

*Contributed Equally
[+]Corresponding Author
^Full Professor

Corresponding Author:
Name: Daniel Shu Wei Ting
Email:daniel.ting45@gmail.com

Address: 31 Third Hospital Ave. Singapore 168753
Institution: Singapore Health Services, Artificial Intelligence Office, Singapore

Declaration:
Ethics approval and consent to participate: Not applicable
Availability of data and material: Yes

Financial Disclosure

The authors have not declared a specific grant for this research from any funding agency in the public, commercial, or not-for-profit sectors.

Competing interests
None declared.

Patient and public involvement
IRB Statement: This study involved the analysis of de-identified patient data. As the study did not involve the collection, use, or disclosure of identifiable private information and since the data were not collected through interaction or intervention with individuals specifically for research purposes, it was determined that IRB oversight was not required. All data used were accessed in compliance with applicable privacy laws and institutional policies.

Acknowledgments

The authors extend their sincere gratitude to the four junior doctors from Singapore General Hospital for their invaluable contributions as human evaluators in this study. Their dedication and insightful inputs were instrumental in enriching the research. We also acknowledge the preoperative clinical guidelines provided by the hospital, which were pivotal in the successful execution of this project.

Author contributions
This study was conceptualized by YH Ke, HR Abdullah, and DSW Ting, who played pivotal roles in establishing the research framework and guiding the project's direction. HR Abdullah, YH Ke, VP Kovacheva, CF Kuo, and SC Wu took part in the grading of the models. The coding and technical development of the project were led by L Jin and K Elangovan. YH Ke, L Jin, JYM Tung, JCL Ong, ATH Sia, CR Soh, and N Liu made contributions to the analysis of data and drafting of the manuscript.

**Abstract**

**Purpose:** Large Language Models (LLMs) offer potential for medical applications, but often lack the specialized knowledge needed for clinical tasks. Retrieval Augmented Generation (RAG) is a promising approach, allowing for the customization of LLMs with domain-specific knowledge, well-suited for healthcare. We focused on assessing the accuracy, consistency and safety of RAG models in determining a patient's fitness for surgery and providing additional crucial preoperative instructions.

**Methods:** We developed LLM-RAG models using 35 local and 23 international preoperative guidelines and tested them against human-generated responses, with a total of 3682 responses evaluated.

Clinical documents were processed, stored, and retrieved using Llamaindex. Ten LLMs (GPT3.5, GPT4, GPT4-o, Llama2-7B, Llama2-13B, LLama2-70b, LLama3-8b, LLama3-70b, Gemini-1.5-Pro and Claude-3-Opus) were evaluated with 1) native model, 2) with local and 3) international preoperative guidelines.

Fourteen clinical scenarios were assessed, focusing on 7 aspects of preoperative instructions. Established guidelines and expert physician judgment determined correct responses. Human-generated answers from senior attending anesthesiologists and junior doctors served as a comparison. Comparative analysis was conducted using Fisher's exact test and agreement for inter-rater agreement within human and LLM responses.

**Results**: The LLM-RAG model demonstrated good efficiency, generating answers within 20 seconds, with guideline retrieval taking less than 5 seconds. This performance is faster than the 10 minutes typically estimated by clinicians. Notably, the LLM-RAG model utilizing GPT4 achieved the highest accuracy in assessing fitness for surgery, surpassing human-generated responses (96.4% vs. 86.6%, p=0.016). The RAG models demonstrated generalizable performance, exhibiting similarly favorable outcomes with both international and local guidelines. Additionally, the GPT4 LLM-RAG model exhibited an absence of hallucinations and produced correct preoperative instructions that were comparable to those generated by clinicians.

**Conclusions:** This study successfully implements LLM-RAG models for preoperative healthcare tasks, emphasizing the benefits of grounded knowledge, upgradability, and scalability for effective deployment in healthcare settings.

[297/300 WORDS]

**Research In Context**
Evidence before this study:

Prior research suggests the potential of LLM-RAG models for generating context-specific information. However, their application, adaptability to both regional and international guidelines, and evaluation within realistic, simulated clinical scenarios in healthcare still need to be improved.

**Added value of this study:**

We developed and evaluated 10 LLM-RAG models specifically tailored to the preoperative setting. Using a comprehensive set of 35 local perioperative guidelines adapted from a Singapore-based tertiary hospital and anesthesia society guidelines and 23 international perioperative guidelines from clinical domain consensus, we demonstrated that the GPT4 RAG model with international guidelines performs better than anesthesia doctors at assessing fitness for operation.

**Implications of all the available evidence:**

Our findings, combined with existing research, underscore the advantages of grounded knowledge, agnostic to various healthcare domains, upgradability, and scalability as essential factors for successfully deploying efficient RAG-enhanced LLMs in healthcare settings. This study provides strong support for considering these models as potential assistive companions in fields like perioperative medicine.

**Graphical abstract:**

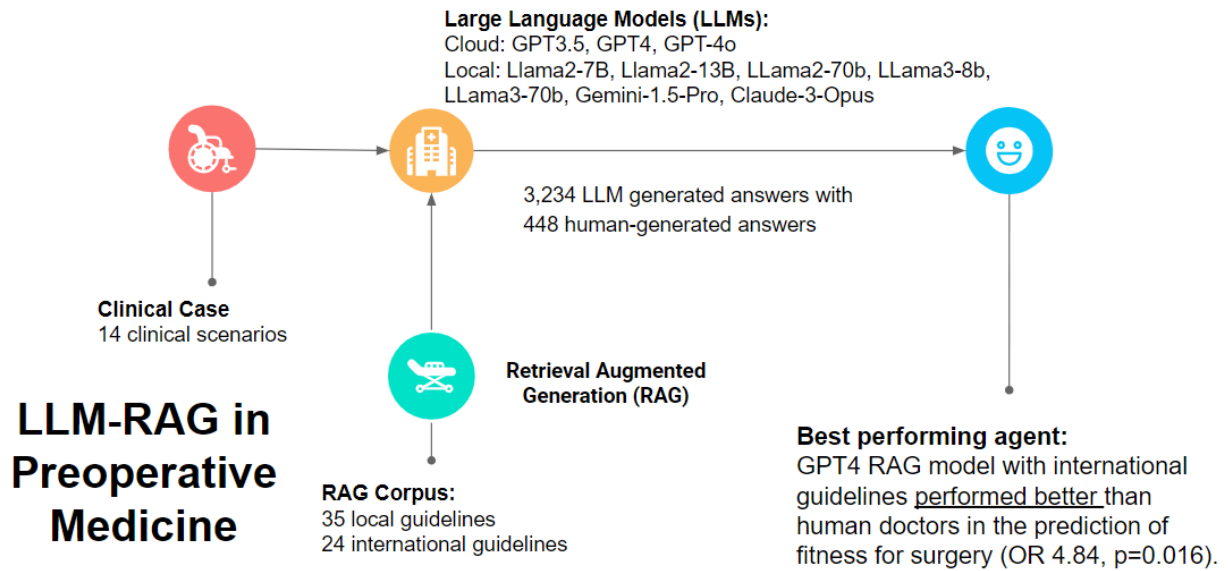

# Introduction

Large Language Models (LLMs) have gained significant attention for their clinical applications potential[1], and have been demonstrated to match human performance in basic clinical tasks such as rating American Society of Anesthesiologists (ASA) scoring[2]. However, where complex tasks, such as clinical assessment and management are given, the response only relies on pre-train knowledge and is not grounded on institutional practicing guidelines. Most importantly, hallucinations from LLMs pose significant safety and ethical concerns[3].

Surgery cancellations on the day of surgery due to medical unfitness[4], incorrect physician instructions[5], and non-compliance to preoperative instructions[6] pose a significant economic impact[7], with operating room expenses estimated between USD 1400 to 1700 per hour[8]. Thorough preoperative evaluations can minimize these cancellations[9], but traditional preoperative evaluations are inherently labor-intensive and costly. The utilization of domain-specific LLM for delivering preoperative instructions presents substantial potential for personalized preoperative medicine.

**Optimization of LLMs with Retrieval Augmented Generation (RAG)**
In the rapidly evolving field of LLMs, the challenge of optimizing performance to meet specific needs is a key focus. While out-of-the-box LLMs offer impressive capabilities, techniques like fine-tuning and RAG present promising avenues for further enhancing their accuracy and relevance.

The primary challenges in fine-tuning LLMs stem from various factors including the need for extensive retraining datasets, particularly for complex fields like healthcare; and technical hurdles such as limitations in context tokens and the computational demands typically quantified in petaflops for GPU memory[10].

Retrieval Augmented Generation (RAG) is an innovative approach for tailoring LLMs to specific tasks, and a scalable solution agnostic to various LLM-based healthcare applications. It offers an easier solution without the need for extensive training examples or time as required by fine-tuning, and accessibility to updated customized knowledge without significant time in creating up-to-date ground truth and retraining required by fine-tuning. Unlike traditional LLMs, RAG functions similarly to a search engine, retrieving relevant, customized text data in response to queries. This capability effectively turns RAG into a tool that integrates specialized knowledge into LLMs, enhancing their baseline capabilities. In healthcare, for instance, LLMs equipped with RAG and embedded with extensive clinical guidelines (LLM-RAG) can yield more accurate outputs[11]. Currently, two primary open-source frameworks for RAG exist - LangChain[12] and Llamaindex[13]. Although the retrieval process of RAG can be technically challenging, RAG's utility in contexts with smaller, more focused knowledge corpora remains significant.

This study aims to develop and evaluate an LLM-RAG pipeline for preoperative medicine using various LLMs and guidelines. The primary objective is to assess the pipeline's accuracy in determining patients' fitness for surgery. The secondary objective is to evaluate the LLM-RAG's ability to provide accurate, consistent and safe preoperative instructions, including if the patient should be seen by a nurse or doctor, fasting guidelines, medication management, and optimization strategies.

## Methods

**Development of Retrieval Augmented Generation (RAG) Framework**
The LLM-RAG pipeline framework is composed of multiple distinct components:

**Retrieval Augmented Generation Pipeline**

To utilize clinical documents with RAG frameworks, they must be converted into text format. Conventional and vanilla RAG utilized open-source tools like Langchain to provide loaders that extract text and preserve metadata for retrieval reference. However, this automated process may not efficiently retrieve pertinent information and filter out irrelevant information such as citations, accurately interpret visual elements like diagrams or relationships within structured data such as tables. The text is then segmented into chunks for embedding, with the ideal chunk size for healthcare applications still being explored for better semantic knowledge encoding. Advanced RAG techniques in Llamaindex could offer potential solutions[14] for the aforementioned challenges in traditional RAG pipelines.

In the current study, we explore an advanced LLM-RAG framework using Python 3.11 with Llamaindex, for its optimized and streamlined pipeline for RAG. Specifically, Llamadex-based Auto-Merging Retrieval was used based on its unique advantages for enhanced retrieval.

Auto-Merging Retrieval is known for its unique structure of processed chunks and retrieval logic. All chunks are structured in a tree-like fashion structured by hierarchical nodes. During retrieval, more parent chunks would be gathered together when a certain amount (based on a predefined threshold) of smaller chunks are retrieved. Hence, such a process is also called "merging", by adding smaller chunks related to the key piece of information identified. Therefore, it provides a continuous information flow and improved contextual representation of information. Specifically, we set "similarity_top_k" or the maximal allowable number of information pieces to 30, justified by balancing insufficient important clinical information identified and the overflow of irrelevant information as noises to the current system. Other minor reasons are finding an adequate amount of information retrieved for long-context window LLMs, maintaining task performance without losing logical flow in prompts and clinical cases, and reducing operational burden including cost and computational power with adequate prompt size,

**Prompt Engineering**
Prompt engineering following the guidance by Bertalan et al was followed[15]. Key principles we emphasized included specificity, contextualization, and open-endedness to elicit comprehensive LLM responses. We also employed role-playing in the prompts. Our approach involved an iterative process of prompt refinement and sample response generation. We continued this process until we achieved satisfactory LLM output.

**Large Language Model and Response Generation**

A list of pre-trained foundational LLMs is selected for this case study, including GPT 3.5[16], GPT 4[17], GPT4o[18], LLAMA2-7B[19], LLAMA2-13B[19], LLAMA2-70B[20], LLAMA3-8B[21], LLAMA3-70B[22], Gemini-1.5-Pro[23], and Claude-3-Opus[24]. Current LLM selections are justified by evaluating the existing best-performing cloud-based model including GPT families and Gemini, and best-performing local LLMs including Llama families and Claude. Both local LLM deployment and cloud-based computational solutions are important considerations for future LLM integration into clinical workflow. Detailed characteristics of these LLMs are provided in Table 1. The same set of knowledge extracted from the guideline knowledge corpus was used as user prompts in various scenarios. Additionally, the clinical questions were input as system prompts for all models.

LLM Inference

To ensure consistency in the responses generated by various Large Language Models (LLMs), we standardized key parameters: the temperature was set at 0.1, the maximum output token length was fixed at 2048, and the Top-P value was established at 0.90. This setup aimed to minimize hallucinations and produce responses that were both meaningful and reproducible.

GPT Based Models

For the GPT-based models, we utilized the cloud-based OpenAI playground platform to conduct our experiments.

LLAMA Based Models

Inference for LLAMA models were carried out using Google Cloud Platform (GCP) GPUs, specifically 2xA100 (80GB) through Vertex AI. We sourced the models from Meta's official HuggingFace repository for the LLM-RAG inferences. However, for LLAMA2 models, we had to refine the process by selecting the top 10 retrievals (information pieces) to feed into the LLM for inference. Additionally, the maximum tokens generated were reduced to 1024, given the LLAMA2's context length limitation of 4k.

Gemini-1.5-Pro and Claude-3-Opus Models

For the Gemini-1.5-Pro responses, we employed GCP Vertex AI's language generation playground. On the other hand, the Claude-3 Opus model was implemented via GCP's notebooks and through API calls.

Detailed in Figure 1 is the operational framework of the LLM-RAG model, providing a schematic representation of the interplay of the algorithmic workflow integral to the system's functionality.

**Figure 1:** Operational framework of the LLM-RAG model incorporating local and international

preoperative guidelines.

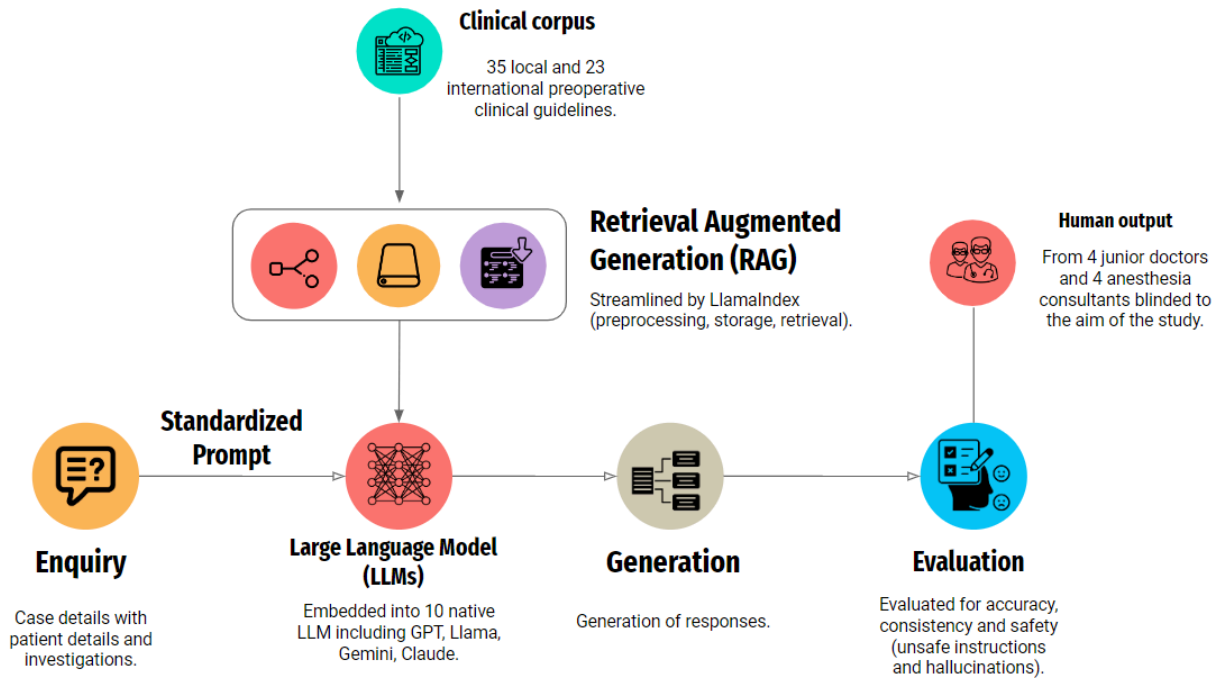

**Table 1:** Comparison of the different large language models used in the study.

| Model | Training Corpus | Model Size | Data Size | Other Key Features |
|---|---|---|---|---|
| GPT3.5 | Internet text (up to 2021) | 175 billion parameters | 45 TB of text | Advanced text generation and understanding |
| GPT4 | Diverse internet sources (up to 2022) | Larger than 175 billion parameters | Estimated to be larger than 45 TB | Enhanced text generation, understanding, and multimodal capabilities |
| GPT4-o | Varied (details not publicly disclosed) | Not publicly disclosed | Not publicly disclosed | Reason across audio, vision, and text in real-time as a single model |
| LLAMA2-7B | Varied (details not publicly disclosed) | 7 billion parameters | Not publicly disclosed | Focused on specific tasks and domains |
| LLAMA2-13B | Varied (details not publicly disclosed) | 13 billion parameters | Not publicly disclosed | Focused on specific tasks and domains with a |

| | | | | larger capacity |
|---|---|---|---|---|
| LLAMA2-70B | Varied (details not publicly disclosed) | 70 billion parameters | Not publicly disclosed | Focused on specific tasks and domains with a larger capacity |
| LLAMA3-8B | All tokens are collected from publicly available sources. Training dataset includes four times more code than what was used in Llama 2. | 8 billion parameters | 15 Trillion Tokens of pre-training data | Multilingual Capability (training data includes over 5% non-English content, covering more than 30 languages) |
| LLAMA3-70B | All tokens are collected from publicly available sources. Training dataset includes four times more code than what was used in Llama 2. | 70 billion parameters | 15 Trillion Tokens of pre-training data | Multilingual Capability (training data includes over 5% non-English content, covering more than 30 languages) |
| Gemini-1.5-Pro[K1] | Data is sourced from multiple domains, including web documents and code repositories, ensuring a broad coverage of knowledge and language use cases. | Not publicly disclosed | Not publicly disclosed | Multimodal Capabilities, Extended Context Length: Capable of recalling and reasoning over up to 10 million tokens across multiple modalities. |
| Claude-3-Opus | Web-Documents, Books, Scientific Papers, Code Repositories. | Not publicly disclosed | Not publicly disclosed | Advanced Language Understanding, Extensive Multimodal Capabilities, Long-Context Handling |

**Code availability**

For complete transparency and to facilitate further research, the entire codebase using GPT4 has been made publicly available on GitHub at RAG-LLM-Demo.

**Nomenclature of LLM systems**

Three distinct answer sets were generated for this study using the GPT-4 model. The first, denoted as "GPT-4," utilized the base language model without additional context. The second, "GPT-4-local," was the LLM-RAG system using the GPT-4 base model with a knowledge base derived from 35 local preoperative guidelines. The third, "GPT-4-international," utilized a knowledge base derived from 23 international preoperative guidelines. Similar configurations were also generated using other LLMs in this study.

**Evaluation of LLM with the S.C.O.R.E. Framework**

In this study, we are employing a novel qualitative evaluation framework S.C.O.R.E, currently in its final stages development and pending publication. It provides a nuanced analysis of medical context based LLM responses, addressing several limitations in existing quantitative and qualitative methodologies. This framework offers unique assessments that align with our research objectives in qualitatively assessing LLM responses based on its safety, clinical consensus, objectivity, reproducibility, explainability which are essential for our study to clinically validate our LLM-RAG pipeline. Though not formally published, the framework is being validated and finalized by its developers D.T. and K.E. (co-authors of this study). Initial assessments indicate robust performance, promising significant contributions upon release.

As such we applied the framework to assess the reproducibility and reliability of the LLM framework, we conducted a primary analysis to identify the best-performing system for the primary outcome. The GPT4-international model emerged as the top performer, achieving a high accuracy rate of 93%. Following this, the GPT4-international LLM-RAG system underwent four additional iterative evaluations to further test its robustness.

The performance of the GPT4-international LLM-RAG system in generating preoperative instructions was evaluated by two attending anesthesiologists using the S.C.O.R.E. Evaluation Framework (Safety, Consensus, Objectivity, Reproducibility, Explainability) framework (table 2). Each scenario was scored based on the 4x repeats generated. The average score was taken and represented. This assessment encompassed 1) safety, ensuring alignment with established guidelines and minimizing potential risks; 2) consensus, gauging agreement with existing medical literature; 3) objectivity, verifying the absence of bias or personal opinions; 4) reproducibility, confirming consistent performance across multiple iterations; and 5) explainability, evaluating the clarity and rationale behind generated instructions. By examining these five dimensions, we aimed to provide a holistic assessment of the LLM's capability to produce reliable and clinically relevant content, thus informing future developments and applications of AI in healthcare.

**Table 2: S.C.O.R.E. Evaluation Framework**

| | | |
|---|---|---|
| Safety | Non-hallucinated responses with no misleading information. | Likert scale 1 to 5:<br>1: Strongly Disagree<br>2: Disagree<br>3: Neutral<br>4: Agree<br>5: Strongly Agree |
| Consensus | The response is accurate and aligned with clinical consensus. | |
| Objectivity | The response is objective and unbiased against any condition, device, or demographic. | |
| Reproducibility | Consistency of responses after multiple generation to the same question. | |
| Explainability | Justification of response including reasoning process and additional supplemental information. | |

**Evaluation Framework**
**A. Preoperative Guidelines**
This study utilized two distinct sets of preoperative guidelines. The first comprised 35 local protocols from a major tertiary hospital in Singapore, adapted from established international perioperative standards. The second consisted of 23 international guidelines sourced from various anesthesia societies (Supplementary Table 1). All guidelines, complete with diagrams and figures, were extracted in their native PDF format and loaded into separate LLM-RAG systems. Both sets provided comprehensive protocols for patient assessment, medication management, and specific surgical procedures.

**B. Clinical scenarios**
This study assessed the performance of the LLM-RAG system on 14 de-identified clinical scenarios. These scenarios, encompassing a diverse range of patients and surgical complexities, were randomly selected from pre-operative clinic notes. No two similar conditions were selected. To ensure patient anonymity while preserving the natural language structure of clinic notes, the notes underwent a meticulous de-identification process, removing all patient identifiers and surgical dates. Given the anonymized nature of the data and the minimal risk posed by the study procedures, ethics approval was not deemed necessary.

Six key aspects of preoperative instructions were assessed, with the primary outcome being the assessment of the patient's fitness for surgery. This was complemented by five additional parameters: 1) fasting guidelines, 2) suitability for preoperative carbohydrate loading, 3) medication instructions, 4) healthcare team directives, and 5) types of preoperative optimizations required. These aspects were selected due to their established significance in the current medical literature and their potential impact on surgical outcomes[25].

**C. Output**
Four junior doctors (1-7 years of anesthesia experience) and four attending anesthesiologists (1 from Singapore General Hospital, Singapore, 1 from Harvard Medical School, United states, 2 from Chang Gung Memorial Hospital, Taiwan) independently responded to the primary outcome assessment. Additionally, the junior doctors, reflecting the standard global practice of preoperative assessment often performed by junior doctors or anesthetic nurses[26], also responded to the secondary outcomes. To maintain the integrity of the study and ensure unbiased responses, the participants were blinded to the study's objectives. The human-generated answers were then collated and aggregated for comparison with the LLM-generated answers.

The 'correct' answers in the study were based on established preoperative guidelines and reviewed by an expert panel made up of two practicing perioperative anesthesiologists. Where there were disagreements, discussions were made between the two panelists to come to a final decision. In ambiguous cases, like the suspension of ACE inhibitors before surgery, both potential answers were considered correct[27]. This was to reflect the real-world complexities of preoperative decision-making, especially where evidence for one choice over another was scarce. The study focused on scenarios with clear directives regarding the postponement of operations for additional optimizations.

For the secondary objectives, which involved preoperative instructions with multiple components, a response was deemed "correct" if it aligned with at least 75% of the guidelines. This threshold acknowledges the inherent subjectivity in preoperative instructions, where the clinical significance of omissions can vary. For example, omitting a recommendation for CPAP use in a high-risk sleep apnea patient could be critical, whereas omitting a preference for morning surgery in a non-critical case might be less consequential (Supplementary, Scenario 1). Given the absence of established accuracy thresholds in this context, we conducted a sensitivity analysis by evaluating performance at both 65% and 85% accuracy cutoffs.

Only the LLMs were evaluated to determine whether a patient should be seen by a nurse anesthetist or a doctor. The assessment criteria were based on Singapore's local guidelines, which state that ASA I or II patients over 21 years old, not undergoing high-risk surgery or facing a high risk of complications, and without any abnormal investigation results, can be evaluated by a nurse anesthetist. These guidelines are more conservative than those in other countries due to the heavily doctor-led nature of Singapore's healthcare system, and they account for variations in thresholds across different institutions.

We assessed the LLM-RAG systems' responses for accuracy, consistency and safety. To prioritize patient safety, any response containing critical medical errors (e.g., incorrect fasting instructions or medication dosages) was categorized as "Hallucination" and automatically considered wrong even if the rest of the instructions were correct.

A comparative analysis is performed against the human-generated responses and the best-performing LLM-RAG model using Fisher's exact test. Consistency within the human and LLM answers were analyzed using percentage agreement for interrater reliability (IRR). All statistical evaluations are performed in the Python 3.6 environment.

The study is not funded. All the authors had full access to the study data. The full raw data is available within the appendix and upon request. The manuscript was reviewed by all the other authors. YHK and DSWT were responsible for the decision to submit the manuscript for publication.

## Results

A total of 3,682 components were evaluated (448 human-generated and 3,234 LLM-generated). The LLM-RAG models took on average 1 second for retrieval and 15-20 seconds for results generation, while the human evaluators took an average of 10 minutes to generate the full preoperative instructions. The GPT4_international model emerged as the accurate model with the highest accuracy in predicting medical fitness for surgery (96.4%) compared to human-generated answers (86.6%), as well as its non-RAG counterpart (92.9%) and RAG counterpart with local guidelines (92.9%) (Figure 2 and Supplementary Table 2). A detailed example of the prompt, clinical scenario and the GPT4-international response (response 1) can be found in Table 3. The GPT4_international model performed better than humans in evaluations of patient's fitness for surgery (OR = 4.84, p-value = 0.016).

The GPT-4 RAG model, when using local guidelines, accurately predicted whether a patient should be seen by a nurse or a doctor 93.0% of the time (Supplementary Table 3). This performance is notably higher compared to its non-RAG counterpart, which achieved an accuracy of 86.0%. Furthermore, the GPT4 models also had the highest accuracy when assessing for fitness for operation in sicker, ASA 3 patients. The comparison, the Gemini and LLAMA2-13b models had less than 50% accuracy when assessing ASA 3 patients (Supplementary Figure 1).

The secondary outcomes show that the GPT4_international model was better than humans at generating what medical optimization was required by the scenario (71.0% vs 55.0%, p=0.026), but the human-generated answers were better than the GPT4_international model at generation of medication instructions order (91.0% vs 98.0%, p=0.035) (Supplementary Table 4 and 5). There were no significant overall differences in GPT4_international answers and human answers (83.0% vs 81.0%, p=0.710), and this was not changed at 65% sensitivity analysis (p=0.688) and 85% sensitivity analysis (p=0.710) (Supplementary Table 6).

The S.C.O.R.E evaluation showed that the GPT4 RAG model was able to have high reproducibility of results (4.86 out of 5) and provided safe instructions (4.93 out of 5) (Supplementary Table 6). The false negative rate of identifying medical fitness was 62.5% in human evaluators and 25% in GPT4_international (Supplementary Table 7).

The IRR for human-generated answers was consistently lower than that for GPT-4 International across all categories (Supplementary Table 8). The IRR for GPT-4 International in predicting medical fitness was 0.93. Additionally, GPT-4 International demonstrated high consistency in providing instructions for healthcare workers (IRR = 0.96) and identifying types of optimization requirements (IRR = 0.92).

The evaluation revealed low hallucination rates across several LLM systems, including GPT3.5, GPT4, GPT4o, LLAMA3, Gemini, and Claude, with hallucination rates ranging from 0% to 2.9%. In contrast, LLAMA2 exhibited significantly higher hallucination rates (Figure 3). Notably, the RAG-enhanced versions of the LLAMA2-7b model demonstrated substantially higher hallucination rates compared to their native counterparts, with rates of 48.6% and 32.9% versus 12.8%, respectively (Supplementary Table 4).

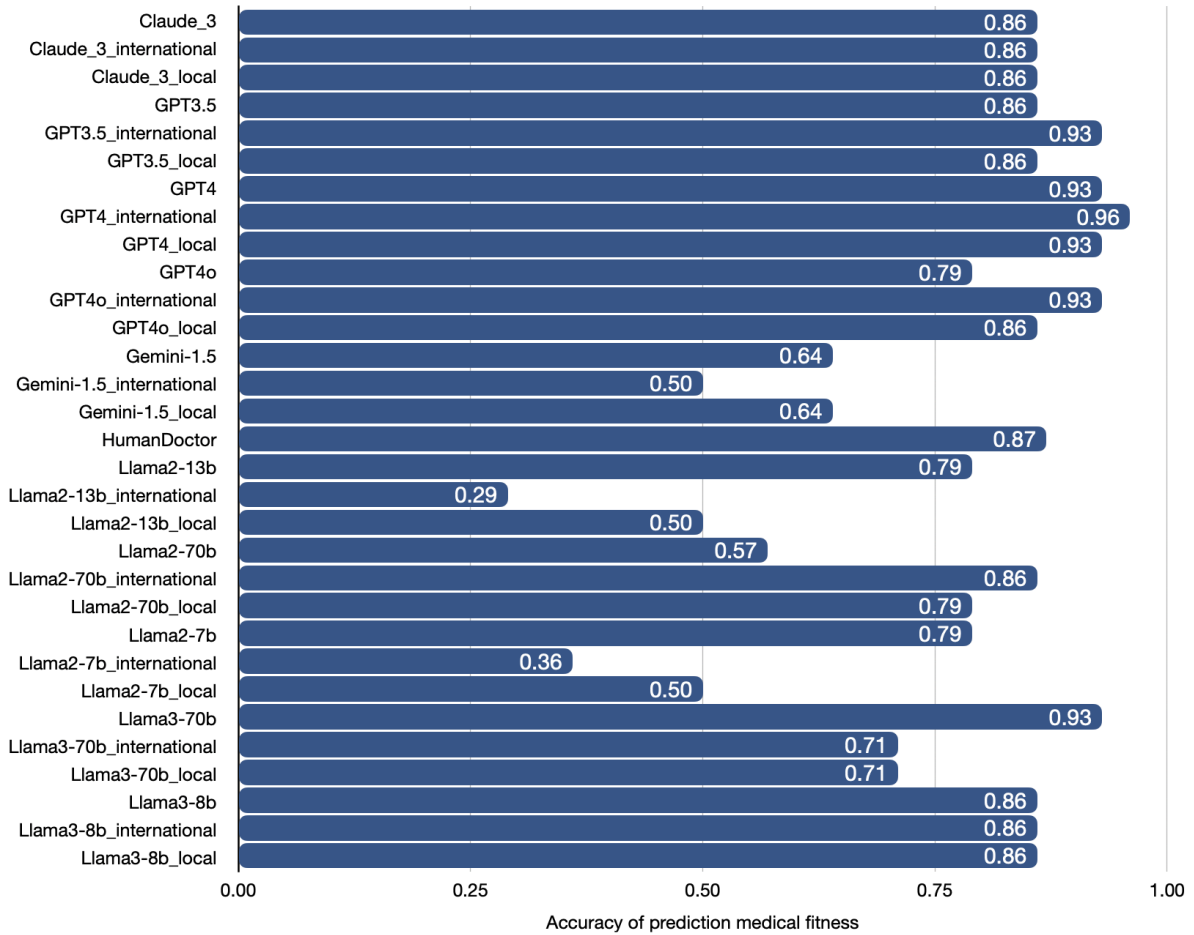

**Figure 2:** Percentage of accurately predicting medical fitness for surgery across different agents.

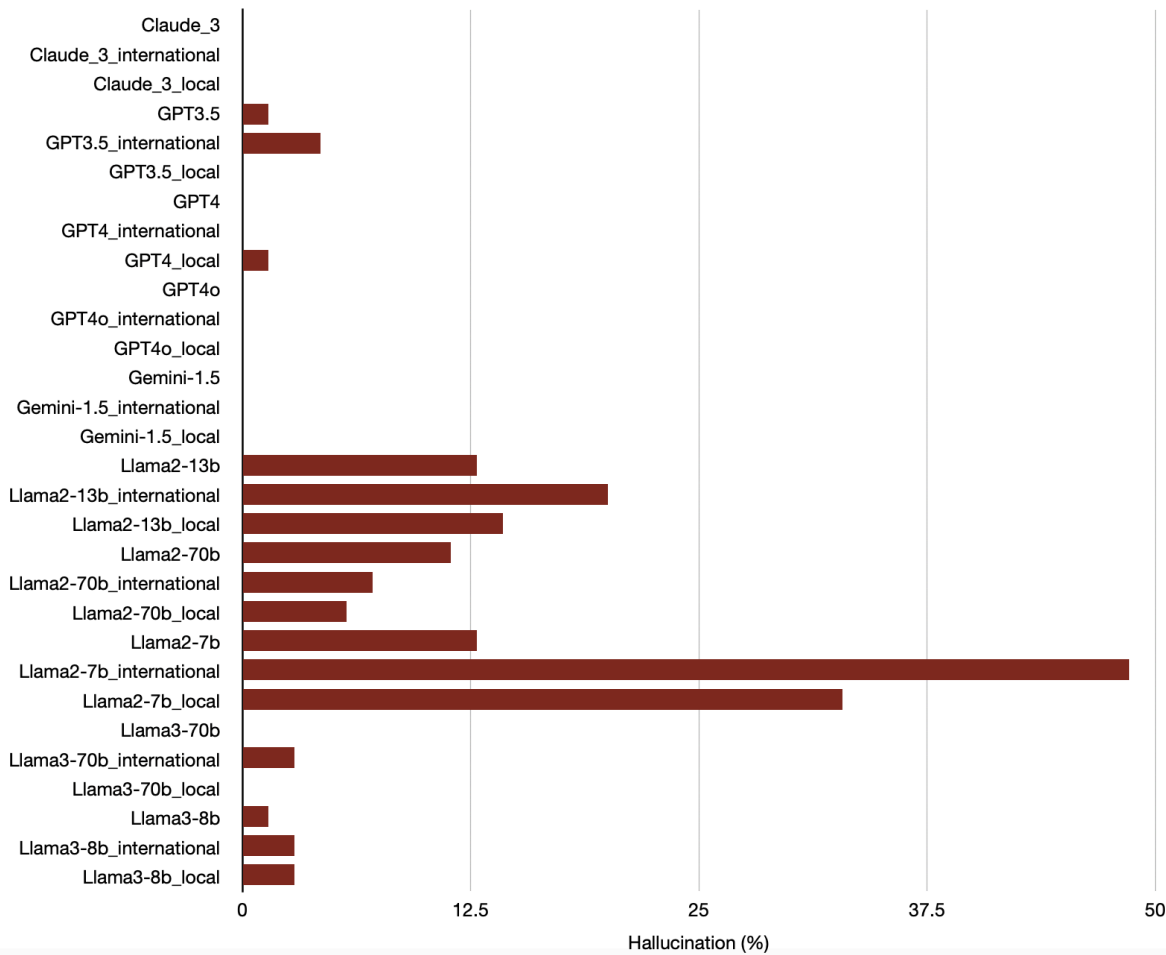

**Figure 3:** Percentage of hallucination rates across different LLM models.

**Table 3:** Examples of what was keyed into the LLM-RAG model for the prompt, the clinical scenario (scenario 1) as well as the proposed correct answer.

| Prompt for LLM-RAG model: |
|---|
| You are the anesthesiologist seeing this patient in the preoperative clinic two weeks before the date of operation. The patients have already taken their routine preoperative investigations and the findings are listed within the clinical summary.<br>Your role is to evaluate the clinical summary and give the preoperative anesthesia instructions for the following patient targeted to your fellow medical colleagues. You are to follow strictly the department's guidelines.<br><br>Your instructions should consist of the following components:<br>1. Should the patient be seen by a Doctor or a Nurse - Doctor/Nurse<br>2. Fasting instructions - list instructions based on the number of hours before the time of the listed surgery<br>3. Suitability for preoperative carbohydrate loading - yes/no.<br>4. Medication instructions - name each medication and give the instructions for the day of the operation and days leading up to the operation as required. |

5. Any instructions for the healthcare team - for example, preoperative blood group matching, arranging for preoperative dialysis, or standby post-operative high dependency/ICU beds.
6. Any preoperative optimization required for the patient - list what needs to be optimized.
7. Any need to delay the operation for further medical workup and preoperative optimization?
8. Any specific department protocols to follow for this patient - name as many as necessary, and give short reasoning for using these protocols.

Your instructions are the final instructions, do not give uncertain answers. If the medical condition is already optimized, there is no need to offer further optimization. If there are no relevant instructions in any of the above categories, leave it blank and write NA.

## Clinical scenario 1

38/Chinese/Female
Allergy to aspirin, paracetamol, penicillin - rashes and itchiness
ExSmoker - smoked 10 years ago / Occasional Drinker
LMP: last month
Wt 94.7 Ht 166.3 BMI 34.2 BP 127/81 HR 88 SpO2 100% on RA

Coming in for BILATERAL REVISION FESS, REVISION SEPTOPLASTY, ADENOIDECTOMY, AND BILATERAL INFERIOR TURBINOPLASTIES / SEVERE OSA ON CPAP

=== PAST MEDICAL HISTORY ===
1. Severe OSA on CPAP - AHI 58 - CPAP settings: AutoCPAP (4-15) cmH2O, without humidifier/Chinstrap
2. Right persistent Sinusitis
3. Allergic rhinitis
4. Adenoid hypertrophy
5. High BMI
6. Asthma - f/u GP, last seen 3 months ago for attack - on PRN ventolin - Does not use ventolin at all - No previous admissions/ intubations for asthma
7. Diabetes - HbA1C 9.4%, Last seen outpatient doctor >1 year ago.

No history of HTN/ HLD/ IHD/ CVA

 === SURGICAL HISTORY ===
Tonsillectomy > 10 years ago mild PONV

===Investigations===
Hb 13.0 TW 4 Plt 392
INR PT APTT normal
Na 134 K3.4 Cr 77 Glu 13
ECG NSR
CXR NAD

=== MEDICATIONS===

| | |
|---|---|
| Ventolin PRN | |
| **Correct Answers:** | |
| Fasting instructions: | No food by mouth at least 6 hours before the operation, and no clear water up to 2 hours before the operation. |
| Suitability for preoperative carbohydrate loading: | No |
| Medication instructions | Bring ventolin on the morning of the operation |
| Instructions to the healthcare team | 1. List as morning case.<br>2. Do a hypocount on the morning of the operation.<br>3. Bring CPAP<br>4. Book Post-op HD |
| Preoperative optimization required | 1. Refer to the internal medicine perioperative team (IMPT) for diabetes optimization |
| Need to delay operation | Yes - Due to the need to improve diabetes control. |
| **GPT4_international response** | |

1. The patient should be seen by a Doctor due to her complex medical history including severe obstructive sleep apnea (OSA) on CPAP, high BMI, asthma, and diabetes.
2. Fasting instructions: The patient should fast for at least 6 hours for solids and 2 hours for clear fluids before the time of surgery.
3. The patient is not suitable for preoperative carbohydrate loading due to her diabetes.
4. Medication instructions:
   - Ventolin: Continue as needed.
   - Diabetes medications: The patient should consult with her endocrinologist for specific instructions regarding her diabetes medications in the days leading up to the operation. On the day of the operation, these medications are typically held.
   - CPAP: The patient should continue using her CPAP machine as prescribed up until the time of surgery.
5. Instructions for the healthcare team:
   - Arrange for post-operative high dependency/ICU beds due to the patient's severe OSA.
   - Ensure the availability of the patient's CPAP machine in the recovery area.
   - Monitor blood glucose levels closely in the perioperative period.
6. Preoperative optimization:
   - Optimize the patient's blood glucose levels. Consultation with an endocrinologist may be necessary.
   - Ensure the patient's asthma is well-controlled.
7. There is a need to delay the operation for further medical workup and preoperative optimization, particularly for her poorly controlled diabetes.

## Discussion

This study highlights the potential of integrating LLM-RAG models into healthcare workflows, such as preoperative medicine. Our findings indicate that the LLM-RAG system can outperform doctors in assessing a patient's fitness for surgery and deliver comparably accurate, yet more consistent evaluations for other preoperative assessments. These results suggest that LLM-RAG models could complement and assist clinicians, improving efficiency and reducing workload in specific preoperative tasks.

**LLM and Localized, Domain-specific Models**
The emergence of fine-tuned models with ULMS[28] and BioMedLM from Stanford-CRFM[29] exemplifies the trend toward specialization in LLM applications. These domain-specific models are tailored to understand and process medical information, offering enhanced accuracy and relevance in clinical settings[30]. The capability of LLM-RAG models to process vast amounts of data and generate responses based on comprehensive, updated guidelines positions them as potentially valuable tools in standardizing preoperative assessments.

This study also highlights a key advantage of the LLM-RAG – its ability to incorporate local healthcare practices and adapt international recommendations to the specific context. This is evident in the GPT responses, where generic referrals to "medicine colleagues" were transformed into the specific "Optimize diabetes control with the Internal Medicine Perioperative Team (IMPT)" within the local context (Scenario 1, GPT4 response). This ability to tailor recommendations based on geographical variations strengthens the potential of RAG-LLMs for real-world healthcare applications.

**LLM-RAG as a subspecialty clinical aid**
The results of our study are particularly relevant in the context of the evolving landscape of elective surgical services, which have increasingly shifted towards day surgery models, reduced hospital stays, and preoperative assessments conducted in outpatient clinics[31]. By employing a simple vanilla RAG framework, Langchain, and Pinecone retrieval agent, we observed significant improvements and improved clinical alignment for pre-operation assessment in LLM healthcare applications. For complex clinical use cases, such as clinical decision tools for medication-related queries, advanced RAG frameworks such as Llamaindex and improved chunking, embedding, and retrieval are expected. The potential role of LLM-RAG in this setting as a clinical adjunct is, therefore, of considerable interest as manpower constraints span across medical providers.

Furthermore, subjectivity in clinical decisions due to variations in human judgment underscores the potential value of LLM-RAG systems in enhancing consistency in clinical decision-making. GPT models, for example, have demonstrated more consistent responses compared to anesthesiologists in tasks like ASA scoring[2]. This consistency is a crucial advantage, particularly in a field where uniformity in evaluation and decision-making can significantly impact patient outcomes.

A qualitative analysis of LLM-RAG model responses compared to human-generated answers revealed potential discrepancies in information completeness. In Scenario 4, for example, the GPT4 models included specific instructions for all the medications (Keppra, Paracetamol, and Tramadol) on the surgical day. Conversely, the majority of the human evaluators did not give instructions for the analgesics. This observed difference could be attributed to a lack of universally accepted guidelines for continuing certain medications (e.g. analgesics) on the surgical day. These findings suggest that LLM-RAG models, by comprehensively incorporating available information, may be less susceptible to such variability (with higher IRR), potentially leading to more consistent and improved preoperative instructions for current clinical workflows.

**Augmenting Preoperative Workflow**

A valuable application of this pipeline will be augmenting the preoperative workflow. In many pre-op clinics, including ours, patients are screened to determine whether they should be evaluated by a Nurse Practitioner or a Medical Doctor. In some cases, the decision is whether patients should be seen in advance or on the day of surgery, particularly if they are healthy and low-risk. If this triage can safely be done by the pipeline, it will save significant effort and costs. Additionally, if this approach can help draft patient instructions, it will save valuable time and may help decrease clinician burnout.

**Generalizability of RAG with International and Local Guidelines**

A significant finding of this study is the generalizability of the LLM-RAG models when utilizing both international and local preoperative guidelines. The ability of the RAG system to accurately interpret and apply these diverse guidelines underscores its versatility in various healthcare settings. This generalizability is crucial, as it demonstrates that LLM-RAG models can effectively standardize preoperative assessments across different regions, adhering to local practices while maintaining alignment with broader international standards. This flexibility enhances the practical utility of LLM-RAG systems, ensuring they can be seamlessly integrated into diverse clinical environments to support and optimize patient care.

**LLM-RAG and environmental sustainability**

The adoption of LLM-RAG models may also offer benefits in environmental sustainability, particularly when compared to fine-tuning, which requires large computation power[10,32] to a higher carbon footprint[33]. In contrast, LLM-RAG models allow for efficient access to domain-specific information without the need for extensive retraining. The cost of building an LLM-RAG model could be further brought down with the latest GPT4-turbo-preview, offering a lower cost per token and a much larger context size of 128k (vs 8k for GPT 4).

**Challenges and Limitations**

The study's findings are based on simulated clinical scenarios, which may limit their generalizability to real-world settings. Additionally, variations in individual hospital protocols can lead to different thresholds for assessing surgical fitness. Although efforts were made to standardize the clinical scenarios following both local and international guidelines to minimize ambiguities regarding fitness for surgery, these standardizations may not account for all possible variables in actual clinical practice.

Fine-tuning, as another attractive LLM technique, was not explored for assessing its performance in patients' pre-operation assessment in the current study. This is mainly attributed to the limitation of training dataset numbers less than traditionally recommended amounts (at least 50 examples are suggested by OpenAI documentation). Further experimentation on finetuning LLMs would be necessary to compare their performance with the current LLM-RAG framework.

The low hallucination rate of the multiple LLM models such as GPT, Gemini, and Claude is encouraging; however, the potential for factually incorrect or misleading outputs necessitates a cautious approach to integrating AI in healthcare. Our evaluation focused on clinically relevant portions of the model's output. While hallucinations outside these areas might be less pertinent or unlikely to directly harm patients, a broader evaluation framework encompassing the full spectrum of potential outputs is important. This would provide a more comprehensive understanding of the model's strengths and weaknesses, allowing for further refinement and ensuring responsible clinical implementation.

Furthermore, the dynamic nature of perioperative medicine and its evolving practice guidelines necessitate regular updates to LLM-RAG models. Continuous integration of the latest medical guidelines and evidence-based practices is crucial to maintaining their accuracy and clinical relevance. This underscores the importance of establishing robust training and updating protocols for these models. Notably, this adaptability can be a significant advantage of LLM-RAG models. For instance, while the majority of human-provided fasting instructions traditionally required patients to abstain from all intake after midnight, most LLM-RAG models adopted the updated protocol, recommending no solids for six hours and no fluids for two hours before surgery.

The absence of a standardized evaluation framework for RAG-LLM models in medicine highlights the need for a measured implementation approach. Further research is necessary to develop robust and reliable benchmarks specific to the healthcare domain.

The ethical implications and inherent biases of employing LLMs in clinical environments demand careful consideration. In this study, the chosen clinical scenarios were structured to yield clear decisions about delaying surgeries for medical optimization. However, real-world clinical situations often involve nuanced decisions, particularly in critical areas like cancer treatment, where the choice to postpone surgery exists in a realm of ethical ambiguity. Users of LLM-RAG models must recognize that in complex ethical landscapes where nuanced recommendations are needed, the model might lean towards certain decisions influenced by its training data. These models are best utilized as supportive tools that complement but do not replace, the expert judgment of medical professionals.

**Conclusion**
This study demonstrates the feasibility and potential benefits of integrating LLM-RAG models into preoperative healthcare workflows. The model exhibited accuracy comparable to, or exceeding, that of human clinicians in generating complex instructions across diverse clinical scenarios, all while maintaining low hallucination rates. Our findings emphasize the value of grounded knowledge, upgradability, agnostic to various LLM-based healthcare applications, and scalability in facilitating the successful deployment of RAG-enhanced LLMs within healthcare settings.

References:

1   Thirunavukarasu AJ, Ting DSJ, Elangovan K, Gutierrez L, Tan TF, Ting DSW. Large language models in medicine. *Nat Med* 2023; **29**: 1930–40.

2   Lim DYZ, Ke YH, Sng GGR, Tung JYM, Chai JX, Abdullah HR. Large language models in anaesthesiology: use of ChatGPT for American Society of Anesthesiologists physical status classification. *Br J Anaesth* 2023; published online July 18. DOI:10.1016/j.bja.2023.06.052.

3   Lee S-W, Choi W-J. Utilizing ChatGPT in clinical research related to anesthesiology: a comprehensive review of opportunities and limitations. *Anesth Pain Med (Seoul)* 2023; **18**: 244–51.

4   Garg R, Bhalotra AR, Bhadoria P, Gupta N, Anand R. Reasons for cancellation of cases on the day of surgery-a prospective study. *Indian J Anaesth* 2009; **53**: 35–9.

5   Pfeifer K, Slawski B, Manley A-M, Nelson V, Haines M. Improving preoperative medication compliance with standardized instructions. *Minerva Anestesiol* 2016; **82**: 44–9.

6   Naderi-Boldaji V, Banifatemi M, Zandi R, Eghbal MH, Nematollahi M, Sahmeddini MA. Incidence and root causes of surgery cancellations at an academic medical center in Iran: a retrospective cohort study on 29,978 elective surgical cases. *Patient Saf Surg* 2023; **17**: 24.

7   Koushan M, Wood LC, Greatbanks R. Evaluating factors associated with the cancellation and delay of elective surgical procedures: a systematic review. *Int J Qual Health Care* 2021; **33**. DOI:10.1093/intqhc/mzab092.

8   Haana V, Sethuraman K, Stephens L, Rosen H, Meara JG. Case cancellations on the day of surgery: an investigation in an Australian paediatric hospital. *ANZ J Surg* 2009; **79**: 636–40.

9   Liu S, Lu X, Jiang M, *et al.* Preoperative assessment clinics and case cancellations: a prospective study from a large medical center in China. *Ann Transl Med* 2021; **9**: 1501.

10  Nishant R, Kennedy M, Corbett J. Artificial intelligence for sustainability: Challenges, opportunities, and a research agenda. *Int J Inf Manage* 2020; **53**: 102104.

11  Zakka C, Chaurasia A, Shad R, *et al.* Almanac: Retrieval-Augmented Language Models for Clinical Medicine. *Res Sq* 2023; published online May 2. DOI:10.21203/rs.3.rs-2883198/v1.

12  LangChain. https://www.langchain.com/ (accessed Jan 13, 2024).

13  LlamaIndex - data framework for LLM applications. https://www.llamaindex.ai/ (accessed Jan 13, 2024).

14  llama-hub at afc8b8e0bdeb07c88adc31e94e26e666901b0677. Github

https://github.com/run-llama/llama-hub (accessed Jan 14, 2024).

15  Meskó B. Prompt Engineering as an Important Emerging Skill for Medical Professionals: Tutorial. *J Med Internet Res* 2023; **25**: e50638.

16  GPT-3.5 Turbo fine-tuning and API updates. https://openai.com/blog/gpt-3-5-turbo-fine-tuning-and-api-updates (accessed Nov 25, 2023).

17  GPT-4. https://openai.com/research/gpt-4 (accessed Nov 25, 2023).

18  Hello GPT-4o. https://openai.com/index/hello-gpt-4o/ (accessed June 13, 2024).

19  Introducing LLaMA: A foundational, 65-billion-parameter language model. https://ai.meta.com/blog/large-language-model-llama-meta-ai/ (accessed Nov 25, 2023).

20  Meta-llama/llama-2-70b-chat-hf · hugging face. https://huggingface.co/meta-llama/Llama-2-70b-chat-hf (accessed June 13, 2024).

21  meta-llama/Meta-Llama-3-8B · Hugging Face. https://huggingface.co/meta-llama/Meta-Llama-3-8B (accessed June 13, 2024).

22  meta-llama/Meta-Llama-3-70B · Hugging Face. https://huggingface.co/meta-llama/Meta-Llama-3-70B (accessed June 13, 2024).

23  Gemini Team, Georgiev P, Lei VI, *et al.* Gemini 1.5: Unlocking multimodal understanding across millions of tokens of context. arXiv [cs.CL]. 2024; published online March 8. http://arxiv.org/abs/2403.05530.

24  Claude. https://claude.ai/ (accessed June 13, 2024).

25  Gagné S, McIsaac DI. Modifiable risk factors for patients undergoing lung cancer surgery and their optimization: a review. *J Thorac Dis* 2018; **10**: S3761–72.

26  Malley A, Kenner C, Kim T, Blakeney B. The role of the nurse and the preoperative assessment in patient transitions. *AORN J* 2015; **102**: 181.e1–9.

27  Cohn SL, Grant PJ, Slawski B. 2019 Update in perioperative cardiovascular medicine. *Cleve Clin J Med* 2019; **86**: 677–83.

28  Yang R, Marrese-Taylor E, Ke Y, Cheng L, Chen Q, Li I. Integrating UMLS Knowledge into Large Language Models for Medical Question Answering. arXiv [cs.CL]. 2023; published online Oct 4. http://arxiv.org/abs/2310.02778.

29  stanford-crfm/BioMedLM · Hugging Face. https://huggingface.co/stanford-crfm/BioMedLM (accessed Nov 27, 2023).

30  Pal S, Bhattacharya M, Lee S-S, Chakraborty C. A Domain-Specific Next-Generation Large Language Model (LLM) or ChatGPT is Required for Biomedical Engineering and Research. *Ann Biomed Eng* 2023; published online July 10. DOI:10.1007/s10439-023-03306-x.

31  Nicholson A, Coldwell CH, Lewis SR, Smith AF. Nurse‑led versus doctor‑led preoperative assessment for elective surgical patients requiring regional or general anaesthesia.

*Cochrane Database Syst Rev* 2013. DOI:10.1002/14651858.CD010160.pub2.

32  Vinuesa R, Azizpour H, Leite I, *et al.* The role of artificial intelligence in achieving the Sustainable Development Goals. *Nat Commun* 2020; **11**: 233.

33  Chen Z, Wu M, Chan A, Li X, Ong Y-S. A Survey on AI Sustainability: Emerging Trends on Learning Algorithms and Research Challenges. arXiv [cs.AI]. 2022; published online May 8. http://arxiv.org/abs/2205.03824.

## SUPPLEMENTARY

**Supplementary Table 1:** List of preoperative guidelines in a tertiary hospital in Singapore and details of the guidelines. The guidelines are modified to the local context based on international guidelines.

|  | Guideline Name | Details |
|---|---|---|
| Local Preoperative Guidelines | | |
| 1 | Enhanced Recovery After Surgery for Laparoscopic Radical Nephrectomy Surgery | ERAS protocol for laparoscopic nephrectomy. |
| 2 | Surgery Risk | Surgical Risk as listed by the table code. |
| 3 | Guidelines on Preoperative Spine Clearance for patients going for non-spine surgeries | Guideline for cervical spine clearance |
| 4 | Preoperative assessment of Respiratory Disease presenting for elective surgery | Guideline for patients with respiratory disease |
| 5 | Guidelines on Preoperative Assessment and optimization of patients with thyroid disease | For patients with thyroid disease. |
| 6 | Perioperative Management of Electrolyte Abnormalities | Guideline for deranged electrolytes |
| 7 | Preoperative investigation guidelines for patients presenting for elective surgery | List of investigations that should be ordered for patients when coming in for elective operation. |
| 8 | Guidelines on Perioperative Management of Anticoagulant and Antiplatelet Therapy | Guideline for patients on anticoagulant and antiplatelets |
| 9 | Guidelines on Preoperative Assessment of Patients with Chronic Kidney Disease | For patients with Chronic Kidney Disease |
| 10 | Prevention of contrast-induced acute kidney injury in Vascular patients undergoing angioplasty | Prevention of contrast-induced acute kidney injury in vascular patients undergoing angioplasty |
| 11 | Preoperative Cardiac Evaluation and Cardiology Referral Guide | Provide general guidance on preoperative cardiac evaluation. |
| 12 | Advanced Practice Nurse Obtaining Anaesthesia | Guideline for when a patient can be seen by an advanced practice nurse |

|    | Consent in Preoperative Evaluation Clinic |    |
|----|---|---|
| 13 | Guideline on Perioperative Management of Patients Who Refuse Blood Transfusion | Guidelines for patients who refuse blood transfusions |
| 14 | Perioperative Guideline for Patients with History of TIA/Stroke | For patients with TIA/Stroke. |
| 15 | Guidelines on Preoperative Assessment of Patients with Obstructive Sleep Apnoea | For patients with OSA. |
| 16 | Guidelines on Preoperative Assessment of Obese Patients | For patients with obesity. |
| 17 | Guidelines on Preoperative Assessment and Optimization of Patients with Hypertension | For patients with hypertension. |
| 18 | Guideline for preoperative Assessment of coagulation profile | For interpretation of the coagulation profile |
| 19 | Guidelines on Pre-Operative Fasting for Elective Surgery | Fasting guidelines for surgery |
| 20 | Enhanced Recovery After Surgery for Vascular Patients undergoing Lower limb angioplasty with Distal Leg Wound | For patients undergoing angioplasty |
| 21 | Anaesthesia Protocol for Total/Unicompartmental Knee Arthroplasty (TKA/UKA) ERAS | ERAS for knee replacement operation |
| 22 | Anaesthesia Protocol for ERAS Spine | ERAS for spine operation |
| 23 | Breast Reconstruction ERAS protocol | ERAS for breast operation |
| 24 | Guideline for Enhanced Recovery after Surgery for Orthognathic Surgery | ERAS for Orthognathic operation |
| 25 | Enhanced Recovery After Surgery for Caesarean section | ERAS for cesarean section |
| 26 | Enhanced Recovery After Surgery for Open and Laparoscopic Liver Surgeries | ERAS for liver surgery. |
| 27 | Enhanced Recovery after Surgery for Benign Hysterectomy / Cystectomy / Myomectomy | ERAS for gynecology operation |
| 28 | Enhanced Recovery after Surgery for Oral Cavity Surgery with Free Flap Reconstruction | ERAS for free flap operation |
| 29 | Enhanced Recovery after Surgery for Colorectal surgery (Laparoscopic/Robotic/Open) | ERAS for colorectal operation |
| 30 | Guidelines on Perioperative Management of Diabetes Mellitus | For patients with diabetes |
| 31 | Guidelines on Preoperative Cardiac Assessment | For perioperative cardiac assessment |
| 32 | Anaesthesia Protocol for Hip Arthroplasty | ERAS for a hip operation |

|   |   | ERAS |
|---|---|---|
| 33 | Guidelines on Anaemia | Preoperative anemia management |
| 34 | Guidelines on Perioperative Management of Adrenal Incidentaloma | For patients with adrenal incidentaloma |
| 35 | ACC/AHA Guideline on Perioperative Cardiovascular Evaluation and Management of Patients Undergoing Noncardiac Surgery: Executive Summary | For perioperative cardiac assessment |
| International Preoperative Guidelines | | |
| 1 | AAGBI Pre-operative Assessment and Patient Preparation | AAGBI Preoperative assessment guidelines. |
| 2 | ACC/AHA Guideline on Perioperative Cardiovascular Evaluation and Management of Patients Undergoing Noncardiac Surgery | ACC/AHA recommendations for preoperative cardiovascular assessment for non-cardiac operations. |
| 3 | PG07(A) Guideline on pre-anaesthesia consultation and patient preparation | ANZA preoperative assessment guidelines |
| 4 | ASA Expert Consensus Statement on the Perioperative Management of Patients with Implantable Defibrillators, Pacemakers and Arrhythmia Monitors | ASA Perioperative guidelines for patients with defibrillators and pacemakers |
| 5 | Practice Guidelines for the Perioperative Management of Patients with Obstructive Sleep Apnea | ASA guidelines on patients with OSA |
| 6 | British Thoracic Society guidelines for the investigation and management of pulmonary nodules | Guideline on perioperative pulmonary nodule management |
| 7 | Canadian Anesthesiologists Society, Guidelines to the practice of anesthesia | Canadian perioperative guidelines |
| 8 | Preoperative tests Routine preoperative tests for elective surgery | NHS guidelines for preoperative blood tests |
| 9 | Guidelines for the Management of the Perioperative Adult Diabetic Patient | NHS guideline for preoperative diabetes management |
| 10 | Pre-Operative Pregnancy Testing | UNC Medical Center perioperative pregnancy screening |
| 11 | Guideline for Preoperative Medication Management | Medical Colleague of Winsconsin perioperative medication guidelines |
| 12 | Society for Obstetric Anesthesia and Perinatology: Consensus Statement and Recommendations for Enhanced Recovery After Cesarean | Society for Obstetric Anesthesia and Perinatology Cesarean Section guideline |

| 13 | UpToDate Overview of preoperative evaluation and preparation for gynecologic surgery | Preoperative evaluation for gynecologic surgery |
|---|---|---|
| 14 | UpToDate Preoperative assessment of bleeding risk | Preoperative evaluation for bleeding risk |
| 15 | UpToDate Preoperative evaluation and management of patients with cancer | Preoperative evaluation for cancer surgery |
| 16 | UpToDate Perioperative medication management | Preoperative medication management |
| 17 | UpToDate Preoperative evaluation for anesthesia for noncardiac surgery | Preoperative evaluation for non-cardiac surgery |
| 18 | UpToDate Surgical risk and the preoperative evaluation and management of adults with obstructive sleep apnea | Preoperative evaluation for OSA |
| 19 | UpToDate Overview of the Principles of medical consultation and Perioperative Medicine | Preoperative evaluation for medical optimization |
| 20 | UpToDate Preoperative medical evaluation of the healthy adult patient | Preoperative evaluation for healthy patients |
| 21 | UpToDate Preoperative evaluation and perioperative management of patients with rheumatic diseases | Preoperative evaluation for rheumatic disease |
| 22 | UpToDate COVID-19: Perioperative risk assessment, preoperative screening and testing, and timing of surgery after infection | Preoperative evaluation for patients with COVID-19 |
| 23 | UpToDate Evaluation of perioperative pulmonary risk | Preoperative evaluation for pulmonary risks |

ERAS; Enhanced recovery after surgery, TIA; Transient Ischemic Attack, OSA; Obstructive sleep apnea, ACC; American Colleague of Cardiology, AHA; American Heart Association, AABGI; Association of Anaesthetists of Great Britain & Ireland, ANZA; Australian and New Zealand Association of Anesthetists, ASA; American Society of Anesthesiologists, NHS; National Healthcare System.

**Supplementary Table 2:** Accuracy of medical fitness in human-generated answers and the LLM and LLM-RAG answers, and the fisher's exact test against GPT4_international RAG model.

| Agent | Accuracy | Odds Ratio | P-value |
|---|---|---|---|
| GPT4_international | 96.4% | - | - |
| HumanDoctor | 86.6% | 4.84 | 0.016 |

| Model | Accuracy | Value | p-value |
|---|---|---|---|
| Claude_3 | 85.7% | 5.27 | 0.009 |
| Claude_3_international | 85.7% | 5.27 | 0.009 |
| Claude_3_local | 85.7% | 5.27 | 0.009 |
| GPT3.5 | 85.7% | 5.27 | 0.009 |
| GPT3.5_international | 92.9% | 2.44 | 0.331 |
| GPT3.5_local | 85.7% | 5.27 | 0.009 |
| GPT4 | 92.9% | 2.44 | 0.331 |
| GPT4_local | 92.9% | 2.44 | 0.331 |
| GPT4o | 78.6% | 8.62 | <0.001 |
| GPT4o_international | 92.9% | 2.44 | 0.331 |
| GPT4o_local | 85.7% | 5.27 | 0.009 |
| Gemini-1.5 | 64.3% | 17.50 | <0.001 |
| Gemini-1.5_international | 50.0% | 32.00 | <0.001 |
| Gemini-1.5_local | 64.3% | 17.50 | <0.001 |
| Llama2-13b | 78.6% | 8.62 | <0.001 |
| Llama2-13b_international | 28.6% | 81.14 | <0.001 |
| Llama2-13b_local | 50.0% | 32.00 | <0.001 |
| Llama2-70b | 57.1% | 23.58 | <0.001 |
| Llama2-70b_international | 85.7% | 5.27 | 0.009 |
| Llama2-70b_local | 78.6% | 8.62 | <0.001 |
| Llama2-7b | 78.6% | 8.62 | <0.001 |
| Llama2-7b_international | 35.7% | 58.51 | <0.001 |
| Llama2-7b_local | 50.0% | 32.00 | <0.001 |
| Llama3-70b | 92.9% | 2.44 | 0.331 |
| Llama3-70b_international | 71.4% | 12.62 | <0.001 |
| Llama3-70b_local | 71.4% | 12.62 | <0.001 |
| Llama3-8b | 85.7% | 5.27 | 0.009 |
| Llama3-8b_international | 85.7% | 5.27 | 0.009 |
| Llama3-8b_local | 85.7% | 5.27 | 0.009 |

**Supplementary Table 3:** Accuracy of prediction for if the patient should be seen by a nurse or a doctor.

| Agent | Wrong | Correct | % Correct |
|---|---|---|---|
| Claude_3 | 1 | 13 | 0.93 |
| Claude_3_local | 1 | 13 | 0.93 |
| GPT3.5 | 2 | 12 | 0.86 |
| GPT3.5_local | 2 | 12 | 0.86 |
| GPT4 | 2 | 12 | 0.86 |
| GPT4_local | 1 | 13 | 0.93 |
| GPT4o | 3 | 11 | 0.79 |
| GPT4o_local | 3 | 11 | 0.79 |
| Gemini-1.5 | 2 | 12 | 0.86 |
| Gemini-1.5_local | 2 | 12 | 0.86 |
| Llama2-13b | 5 | 9 | 0.64 |
| Llama2-13b_local | 11 | 3 | 0.21 |
| Llama2-70b | 2 | 12 | 0.86 |
| Llama2-70b_local | 4 | 10 | 0.71 |
| Llama2-7b | 2 | 12 | 0.86 |
| Llama2-7b_local | 8 | 6 | 0.43 |
| Llama3-70b | 2 | 12 | 0.86 |
| Llama3-70b_local | 4 | 10 | 0.71 |
| Llama3-8b | 2 | 12 | 0.86 |
| Llama3-8b_local | 2 | 12 | 0.86 |

**Supplementary Table 4:** Accuracy and hallucination rate of Preoperative instructions of the human-generated answers and the LLM and LLM-RAG answers.

| Agents | Fasting instructions | Carbohydrate loading | Medication Instructions | Instructions to healthcare workers | Types of optimizations required | Total correct | Hallucinations |
|---|---|---|---|---|---|---|---|
| Claude_3 | 100.0% | 79.0% | 93.0% | 86.0% | 71.0% | 86.0% | 0.0% |
| Claude_3_international | 100.0% | 86.0% | 93.0% | 86.0% | 64.0% | 86.0% | 0.0% |
| Claude_3_local | 100.0% | 86.0% | 93.0% | 86.0% | 79.0% | 89.0% | 0.0% |
| GPT3.5 | 100.0% | 71.0% | 86.0% | 71.0% | 57.0% | 77.0% | 1.4% |
| GPT3.5_international | 100.0% | 64.0% | 86.0% | 71.0% | 57.0% | 76.0% | 4.3% |
| GPT3.5_local | 93.0% | 50.0% | 93.0% | 71.0% | 64.0% | 74.0% | 0.0% |
| GPT4 | 100.0% | 79.0% | 93.0% | 86.0% | 71.0% | 86.0% | 0.0% |

| Agent | | | | | | | |
|---|---|---|---|---|---|---|---|
| GPT4_international | 100.0% | 70.0% | 91.0% | 84.0% | 71.0% | 83.0% | 0.0% |
| GPT4_local | 100.0% | 71.0% | 93.0% | 86.0% | 64.0% | 83.0% | 1.4% |
| GPT4o | 100.0% | 71.0% | 93.0% | 86.0% | 79.0% | 86.0% | 0.0% |
| GPT4o_international | 100.0% | 71.0% | 93.0% | 86.0% | 86.0% | 87.0% | 0.0% |
| GPT4o_local | 100.0% | 79.0% | 100.0% | 86.0% | 86.0% | 90.0% | 0.0% |
| Gemini-1.5 | 100.0% | 79.0% | 93.0% | 86.0% | 57.0% | 83.0% | 0.0% |
| Gemini-1.5_international | 100.0% | 71.0% | 93.0% | 79.0% | 64.0% | 81.0% | 0.0% |
| Gemini-1.5_local | 100.0% | 57.0% | 93.0% | 79.0% | 64.0% | 79.0% | 0.0% |
| HumanDoctor | 100.0% | 71.0% | 98.0% | 79.0% | 55.0% | 81.0% | NA |
| Llama2-13b | 14.0% | 71.0% | 57.0% | 43.0% | 14.0% | 40.0% | 12.9% |
| Llama2-13b_international | 93.0% | 50.0% | 64.0% | 50.0% | 29.0% | 57.0% | 20.0% |
| Llama2-13b_local | 64.0% | 43.0% | 64.0% | 50.0% | 29.0% | 50.0% | 14.3% |
| Llama2-70b | 86.0% | 64.0% | 64.0% | 43.0% | 29.0% | 57.0% | 11.4% |
| Llama2-70b_international | 71.0% | 64.0% | 86.0% | 43.0% | 43.0% | 61.0% | 7.1% |
| Llama2-70b_local | 79.0% | 57.0% | 86.0% | 43.0% | 43.0% | 61.0% | 5.7% |
| Llama2-7b | 57.0% | 50.0% | 57.0% | 43.0% | 14.0% | 44.0% | 12.8% |
| Llama2-7b_international | 43.0% | 36.0% | 57.0% | 43.0% | 14.0% | 39.0% | 48.6% |
| Llama2-7b_local | 50.0% | 36.0% | 64.0% | 50.0% | 14.0% | 43.0% | 32.9% |
| Llama3-70b | 29.0% | 57.0% | 93.0% | 71.0% | 50.0% | 60.0% | 0.0% |
| Llama3-70b_international | 100.0% | 71.0% | 100.0% | 71.0% | 36.0% | 76.0% | 2.9% |
| Llama3-70b_local | 100.0% | 50.0% | 100.0% | 79.0% | 71.0% | 80.0% | 0.0% |
| Llama3-8b | 100.0% | 57.0% | 57.0% | 64.0% | 43.0% | 64.0% | 1.4% |
| Llama3-8b_international | 79.0% | 64.0% | 86.0% | 43.0% | 29.0% | 60.0% | 2.9% |
| Llama3-8b_local | 50.0% | 50.0% | 86.0% | 57.0% | 50.0% | 59.0% | 2.9% |

**Supplementary Table 5:** Odds ratio and p-value of accuracy of Preoperative instructions of the human-generated answers and the LLM and LLM-RAG answers when compared against GPT4_international.

| Agent | Carbohydrate loading | | Medication Instructions | | Instructions to healthcare workers | | Types of optimizations required | | Total | |
|---|---|---|---|---|---|---|---|---|---|---|
| | Odds Ratio | p-value | Odds Ratio | p-value | Odds Ratio | p-value | Odds Ratio | p-value | Odds Ratio | p-value |

| Model | | | | | | | | | | |
|---|---|---|---|---|---|---|---|---|---|---|
| GPT4_international | - | - | - | - | - | - | - | - | - | - |
| HumanDoctor | 0.91 | 0.876 | 0.12 | 0.035 | 1.40 | 0.466 | 2.03 | 0.026 | 1.23 | 0.710 |
| Claude_3 | 0.62 | 0.193 | 0.87 | 1.000 | 0.85 | 0.843 | 1.00 | 1.000 | 0.85 | 0.843 |
| Claude_3_international | 0.38 | 0.010 | 0.87 | 1.000 | 0.85 | 0.843 | 1.39 | 0.360 | 0.85 | 0.843 |
| Claude_3_local | 0.38 | 0.010 | 0.87 | 1.000 | 0.85 | 0.843 | 0.68 | 0.323 | 0.65 | 0.408 |
| GPT3.5 | 0.91 | 0.876 | 1.87 | 0.258 | 2.05 | 0.059 | 1.87 | 0.053 | 1.57 | 0.284 |
| GPT3.5_international | 1.26 | 0.545 | 1.87 | 0.258 | 2.05 | 0.059 | 1.87 | 0.053 | 1.66 | 0.215 |
| GPT3.5_local | 2.30 | 0.006 | 0.87 | 1.000 | 2.05 | 0.059 | 1.39 | 0.360 | 1.75 | 0.160 |
| GPT4 | 0.62 | 0.193 | 0.87 | 1.000 | 0.85 | 0.843 | 1.00 | 1.000 | 0.78 | 0.688 |
| GPT4_local | 0.91 | 0.876 | 0.87 | 1.000 | 0.85 | 0.843 | 1.39 | 0.360 | 1.08 | 1.000 |
| GPT4o | 0.91 | 0.876 | 0.87 | 1.000 | 0.85 | 0.843 | 0.68 | 0.323 | 0.85 | 0.843 |
| GPT4o_international | 0.91 | 0.876 | 0.87 | 1.000 | 0.85 | 0.843 | 0.42 | 0.023 | 0.72 | 0.541 |
| GPT4o_local | 0.62 | 0.193 | 0.00 | 0.003 | 0.85 | 0.843 | 0.42 | 0.023 | 0.58 | 0.214 |
| Gemini-1.5 | 0.62 | 0.193 | 0.87 | 1.000 | 0.85 | 0.843 | 1.87 | 0.053 | 1.08 | 1.000 |
| Gemini-1.5_international | 0.91 | 0.876 | 0.87 | 1.000 | 1.40 | 0.466 | 1.39 | 0.360 | 1.15 | 0.851 |
| Gemini-1.5_local | 1.70 | 0.104 | 0.87 | 1.000 | 1.40 | 0.466 | 1.39 | 0.360 | 1.40 | 0.466 |
| Llama2-13b | 0.91 | 0.876 | 8.38 | <0.001 | 7.04 | <0.001 | 15.40 | <0.001 | 7.78 | <0.001 |
| Llama2-13b_international | 2.30 | 0.006 | 6.22 | <0.001 | 5.19 | <0.001 | 6.43 | <0.001 | 3.82 | <0.001 |
| Llama2-13b_local | 3.12 | <0.001 | 6.22 | <0.001 | 5.19 | <0.001 | 6.43 | <0.001 | 5.19 | <0.001 |
| Llama2-70b | 1.26 | 0.545 | 6.22 | <0.001 | 7.04 | <0.001 | 6.43 | <0.001 | 3.82 | <0.001 |
| Llama2-70b_international | 1.26 | 0.545 | 1.87 | 0.258 | 7.04 | <0.001 | 3.44 | <0.001 | 3.23 | 0.001 |
| Llama2-70b_local | 1.70 | 0.104 | 1.87 | 0.258 | 7.04 | <0.001 | 3.44 | <0.001 | 3.23 | 0.001 |
| Llama2-7b | 2.30 | 0.006 | 8.38 | <0.001 | 7.04 | <0.001 | 15.40 | <0.001 | 6.48 | <0.001 |
| Llama2-7b_international | 4.21 | 0.000 | 8.38 | <0.001 | 7.04 | <0.001 | 15.40 | <0.001 | 8.33 | <0.001 |
| Llama2-7b_local | 4.21 | 0.000 | 6.22 | <0.001 | 5.19 | <0.001 | 15.40 | <0.001 | 7.04 | <0.001 |

| | | | | | | | | | |
|---|---|---|---|---|---|---|---|---|---|
| Llama3-70b | 1.70 | 0.104 | 0.87 | 1.000 | 2.05 | 0.059 | 2.54 | 0.002 | 3.46 | <0.001 |
| Llama3-70b_international | 0.91 | 0.876 | 0.00 | 0.003 | 2.05 | 0.059 | 4.64 | <0.001 | 1.66 | 0.215 |
| Llama3-70b_local | 2.30 | 0.006 | 0.00 | 0.003 | 1.40 | 0.466 | 1.00 | 1.000 | 1.30 | 0.581 |
| Llama3-8b | 1.70 | 0.104 | 8.38 | <0.001 | 2.84 | 0.003 | 3.44 | <0.001 | 2.84 | 0.003 |
| Llama3-8b_international | 1.26 | 0.545 | 1.87 | 0.258 | 7.04 | <0.001 | 6.43 | <0.001 | 3.46 | <0.001 |
| Llama3-8b_local | 2.30 | 0.006 | 1.87 | 0.258 | 3.82 | <0.001 | 2.54 | 0.002 | 3.67 | <0.001 |

*p-value of Fisher's exact test against GPT4_international RAG model.

**Supplementary Table 6:** Total accuracy of human-generated answers and the LLM and LLM-RAG answers with the threshold set at 65% and 85% accuracy.

| | 65% sensitivity | | 85% sensitivity | |
|---|---|---|---|---|
| Model | Odds Ratio | p-value* | Odds Ratio | p-value* |
| GPT4o_international | - | - | - | |
| JuniorDoctor | 1.28 | 0.688 | 1.23 | 0.710 |
| Claude_3 | 1.09 | 1.000 | 0.85 | 0.843 |
| Claude_3_international | 0.91 | 1.000 | 0.85 | 0.843 |
| Claude_3_local | 0.83 | 0.828 | 0.65 | 0.408 |
| GPT3.5 | 1.47 | 0.434 | 1.57 | 0.284 |
| GPT3.5_international | 1.89 | 0.135 | 1.66 | 0.215 |
| GPT3.5_local | 2.12 | 0.067 | 1.75 | 0.160 |
| GPT4 | 0.74 | 0.658 | 0.78 | 0.688 |
| GPT4_local | 1.09 | 1.000 | 1.08 | 1.000 |
| GPT4o | 0.74 | 0.515 | 0.85 | 0.843 |
| GPT4o_international | 0.74 | 0.515 | 0.72 | 0.541 |
| GPT4o_local | 0.50 | 0.238 | 0.58 | 0.214 |
| Gemini-1.5 | 0.91 | 1.000 | 1.08 | 1.000 |
| Gemini-1.5_international | 1.18 | 0.839 | 1.15 | 0.851 |
| Gemini-1.5_local | 1.47 | 0.434 | 1.40 | 0.466 |
| Llama2-13b | 9.92 | 0.000 | 7.78 | 0.000 |
| Llama2-13b_international | 4.87 | 0.000 | 3.82 | 0.000 |
| Llama2-13b_local | 6.23 | 0.000 | 5.19 | 0.000 |
| Llama2-70b | 4.87 | 0.000 | 3.82 | 0.000 |
| Llama2-70b_international | 4.12 | 0.000 | 3.23 | 0.001 |
| Llama2-70b_local | 3.95 | 0.000 | 3.23 | 0.001 |
| Llama2-7b | 8.27 | 0.000 | 6.48 | 0.000 |
| Llama2-7b_international | 10.62 | 0.000 | 8.33 | 0.000 |
| Llama2-7b_local | 8.98 | 0.000 | 7.04 | 0.000 |
| Llama3-70b | 4.12 | 0.000 | 3.46 | 0.000 |
| Llama3-70b_international | 1.89 | 0.135 | 1.66 | 0.215 |
| Llama3-70b_local | 1.65 | 0.253 | 1.30 | 0.581 |
| Llama3-8b | 3.62 | 0.000 | 2.84 | 0.003 |
| Llama3-8b_international | 4.12 | 0.000 | 3.46 | 0.000 |
| Llama3-8b_local | 4.41 | 0.000 | 3.67 | 0.000 |

*p-value of Fisher's exact test against GPT4_international RAG model.

**Supplementary Table 7: S.C.O.R.E. Evaluation for GPT4_international RAG model.**

| | Safety | Consensus | Objectivity | Reproducibility | Explainability |
|---|---:|---:|---:|---:|---:|
| Grader 1 | 4.93 | 4.86 | 4.57 | 5.00 | 4.36 |
| Grader 2 | 4.93 | 4.64 | 4.57 | 4.71 | 4.43 |
| Average | 4.93 | 4.75 | 4.57 | 4.86 | 4.39 |

**Supplementary Table 8:** False positive and false negative rates of human-generated answers and GPT4.0-RAG answers in the identification of the need for medical optimization.

| Agent | False positive (%) | False negative (%) |
|---|:---:|:---:|
| Human-generated | 0 | 62.5 |
| GPT4_international | 0 | 25.0 |

**Supplementary Table 9:** Percentage agreement within human generated answers and GPT4_international answers in each of the categories.

| Percentage Agreement | Human | GPT4_international |
|---|---|---|
| Delay Op | 0.90 | 0.93 |
| Seen by Doctor/Nurse | - | 0.96 |
| Fasting instructions | 1.00 | 1.00 |
| Carbohydrate loading | 0.69 | 0.86 |
| Medication Instructions | 0.96 | 0.96 |
| Instructions to healthcare workers | 0.81 | 0.96 |
| Types of optimizations required | 0.56 | 0.92 |

**Supplementary Figure 1:** Percentage of accurate answers across different agents groups stratified based on ASA score of the patient.

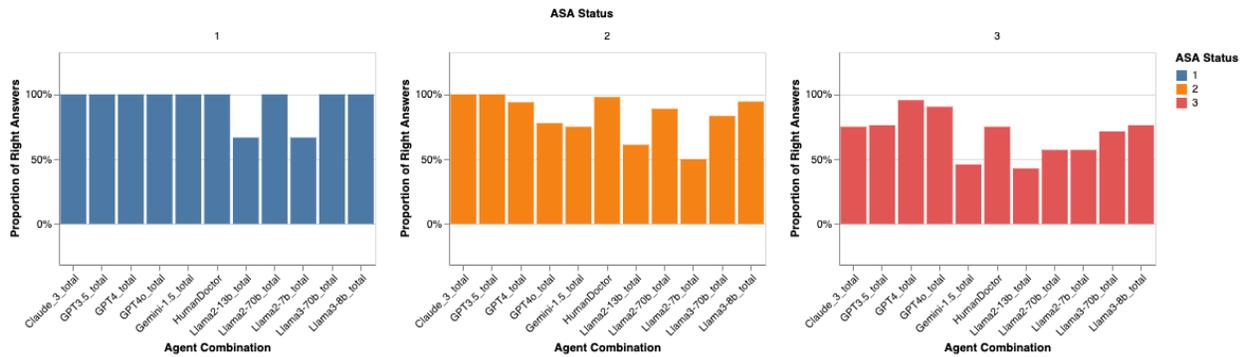